# Racing to Learn: Statistical Inference and Learning in a Single Spiking Neuron with Adaptive Kernels

Saeed Afshar[1], Libin George[2], Jonathan Tapson[1], André van Schaik[1], Tara Julia Hamilton[1,2]

[1]Bioelectronics and Neurosciences, The MARCS Institute, University of Western Sydney, Penrith NSW Australia
[2]School of Electrical Engineering and Telecommunications, The University of New South Wales, Sydney NSW Australia

Correspondence:
Saeed Afshar
The MARCS Institute
Bioelectronics and Neuroscience
University of Western Sydney
Locked Bag 1797
Penrith NSW 2751
Australia
s.afshar@uws.edu.au

## Abstract

This paper describes the Synapto-dendritic Kernel Adapting Neuron (SKAN), a simple spiking neuron model that performs statistical inference and unsupervised learning of spatiotemporal spike patterns. SKAN is the first proposed neuron model to investigate the effects of dynamic synapto-dendritic kernels and demonstrate their computational power even at the single neuron scale. The rule-set defining the neuron is simple: there are no complex mathematical operations such as normalization, exponentiation or even multiplication. The functionalities of SKAN emerge from the real-time interaction of simple additive and binary processes. Like a biological neuron, SKAN is robust to signal and parameter noise, and can utilize both in its operations. At the network scale neurons are locked in a race with each other with the fastest neuron to spike effectively 'hiding' its learnt pattern from its neighbors. The robustness to noise, high speed and simple building blocks not only make SKAN an interesting neuron model in computational neuroscience, but also make it ideal for implementation in digital and analog neuromorphic systems which is demonstrated through an implementation in a Field Programmable Gate Array (FPGA).

Matlab, Python and Verilog implementations of SKAN are available at:
http://www.uws.edu.au/bioelectronics_neuroscience/bens/reproducible_research

## Introduction

### Prior work

Real neurons, the electrically excitable cells of the Eumetazoan, constitute an extremely diverse intractably complex community whose dynamic structures and functions defy all but the broadest generalizations [1][2]. In order to minimize this complexity, the field of Artificial Neural Networks (ANN) has traditionally modeled neurons as deterministic, centrally clocked elements which operate on real valued signals [3]. These signals represent neuronal rate coding where the spiking rate of a neuron encodes useful information and the adjustment of synaptic weights results in learning. This scheme, while mathematically amenable incurs a significant energy cost by discarding the rich temporal information available in the real signals used by neurons to communicate [4][5][6]. In contrast, the highly optimized, low power, portable signal processing and control system that is the brain readily uses temporal information embedded in the input signals and internal dynamics of its stochastic heterogeneous elements to process information [7].

More recently, the greater efficiency, higher performance and biologically realistic dynamics of temporal coding neural networks has motivated the development of synaptic weight adaptation schemes that operate on temporally coding Spiking Neural Networks (SNN) [8][9][10][11][12][13]. After proposition many of these models are followed soon by their implementation in neuromorphic hardware [14][15][16][17][18][19]. One of the problems faced by neuromorphic hardware engineers is the hardware inefficiency of many neural network algorithms. These algorithms are almost always initially designed for performance in a constraint free mathematical context with numerous all-to-all connected neurons and/or to satisfy some biological realism criteria, which create difficulties in hardware implementation.

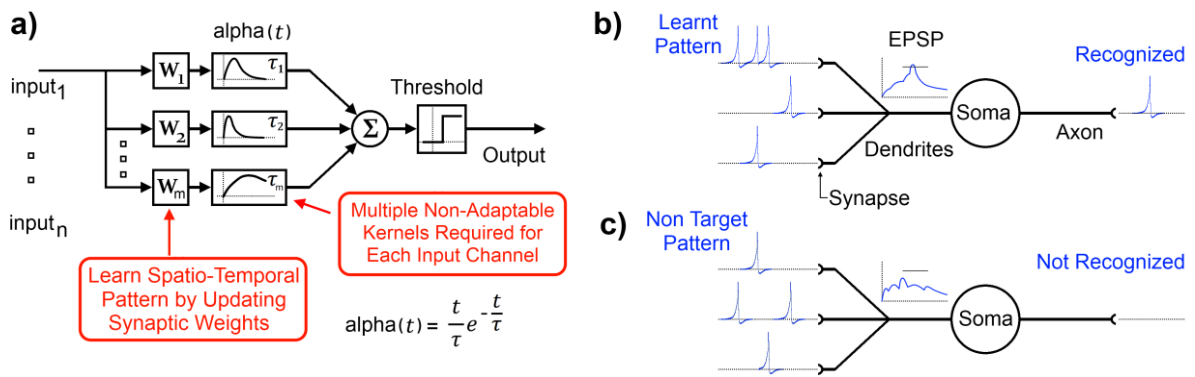

Figure 1. a) Typical functional model of a spiking neuron with static synaptic transfer functions that provide memory of recent spikes. b) Biological representation of the neuron showing the learnt input spike pattern, the resultant Excitatory Post-Synaptic Potentiation (EPSP) and the output spike indicating pattern recognition. c) Presentation of a non-target pattern results in an EPSP that does not cross the threshold producing no output spike.

Additionally in order for such spiking systems to combine temporal coding and weight adaptation, multiple synapses consisting of synaptic transfer functions (or synaptic kernels) as well as synaptic weights must be realized for every input channel as shown in Figure 1. With the aim of being biologically plausible, exponentially decaying functions are typically chosen as the synaptic kernel, which is then multiplied by the synaptic weight. Such functions and weights are quite complex and difficult to implement in simple scalable analog and digital hardware with even the simplest schemes requiring at least one multiplication operation at every synapse. The difficulty of realizing multipliers at the synapse and the large number of synapses used in most algorithms has motivated moves toward more scalable digital synapses [20][21][22][23], novel memristor based solutions [24][25] and second order solutions such as sparse coding [26], time multiplexing, and Address Event Representation (AER) [27] where only one or a few instances of the complex computational units are realized and these are utilized serially. Despite the success of these approaches such serial implementations can sometimes introduce associated bottlenecks, which can detract from the main strength of the neural network approach: its *distributed* nature [28].

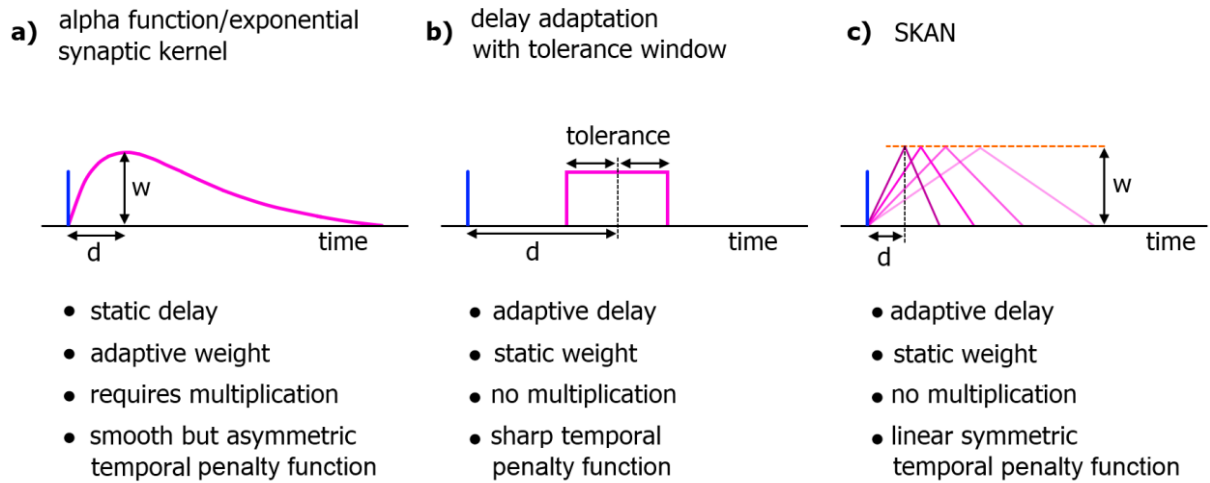

**Figure 2 Comparison of neuromorphic implementations of synapto-dendritic kernels.** The characteristics of realized EPSP kernels are computationally important just prior to their being summed at the soma. These kernels represent the penalty function used to translate the temporal error in spatiotemporal spike patterns at the synapse to the membrane potential at the soma. Due to their large numbers, the complexity, functionality and hardware cost of these kernels are a critical feature of neuromorphic spiking networks.

Rather than implement complex synaptic weight adaptation, other neuromorphic SNN implementations have, in the last three years, focused exclusively on adjustment of explicit propagation delays along the neural signal path and coincidence detection of input spikes to encode memory [29][30][31][32][33][34]. This discarding of synaptic weights and kernels significantly simplifies implementation and improves scalability. The disadvantage is that explicit delay learning schemes can produce "sharp" systems with poor tolerance for the dynamically changing temporal variance they inevitably encounter in applications where neuronal systems are expected to excel: noisy, dynamic and unpredictable environments.

One of the features shared by all the preceding systems is that the kernels used for encoding temporal information are static as shown in Figure 2. However recent advances in neurophysiology have revealed that synapto-dendritic structures and their associated transfer functions are highly complex and adapt during learning in response to the statistical contexts of their stimulus environment [35][36][37][38][39][40][41]. These discoveries are significant in the context of the computational power of even single biological neurons. Whereas in the traditional neuron model synapto-dendritic structures function as weights and cables connecting one soma to the next, the recent findings have demonstrated a wide range of signal integration and processing occurring along the signal path, which confers considerable computational power to single neurons [42][43][44][45]. These effects represent novel dynamics with as yet unexplored emergent computational properties, which may potentially solve currently intractable problems in computational neuroscience [46][47]. These dendritic adaptation effects have recently been modelled through large rule sets [48][49][50] and in the neuromorphic field the use of dendrites for computation is beginning to be explored [51][52][53][54]. However with biological realism as a major focus, many of the models carry significant extra complexity which can impede scalibility.

## Neurons as functional models of distributed processing

In this paper the goal of performance in hardware motivates a change in focus from claims of accurate modeling of computation in biological neurons to exploiting the computational power of artificial but biologically inspired neurons. These are herein defined as a set of simple distributed informational processing units that communicate through binary valued pulses (spikes), receive inputs from multiple input channels (synapses and dendrites), and have a single output channel (axon).

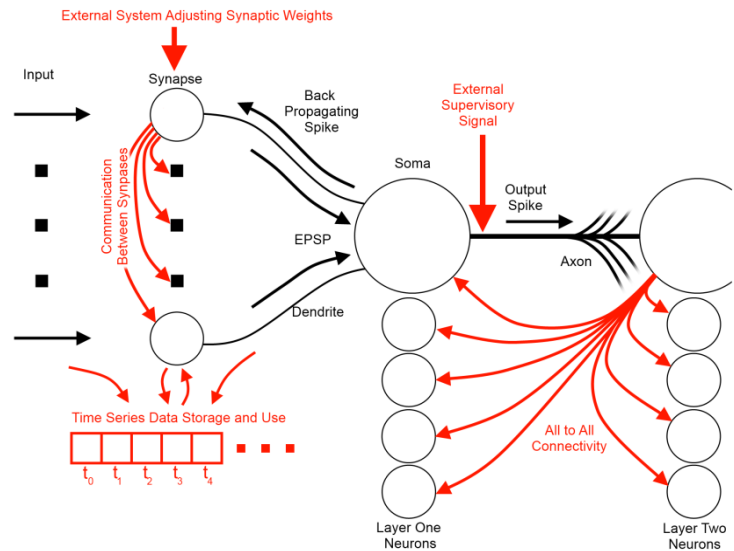

**Figure 3. Information flow schemes in unsupervised spiking neural network algorithms and their impact on hardware implementation.** Black indicates the fundamental elements and information paths of a spiking neural network. Red indicates added features and information paths that can cause difficulties in hardware, or limit algorithm utility.

Figure 3 illustrates the basic elements of spiking neural network algorithms as well as some useful information flow and storage restrictions (red), which, if adhered to at the neuron design stage, prove helpful during the physical implementation stage. These restrictions include:

1. Self-contained: In a self-contained system, no external controlling system is required for the system to function. Examples of systems that are *not* self-contained include synapses that require adjustment via an external controller, or systems that assume an external supervisor in real world contexts where such a signal is unlikely to be available.
2. Scalable connectivity: Systems that require all-to-all connectivity between the neurons or where the synapses or dendrites directly communicate their weights or potentials to each other are not hardware scalable or biologically possible. All-to-all connected neurons require a geometrically increasing number of connections, which is prohibitive both in hardware and in the brain [55][56].
3. Storage of time series data: Systems whose processing units require large segments of their time series data to be stored and be accessible for later processing in the fashion of standard processors require a significant amount of on-site memory not possible in biological systems and would add significant complexity to neuromorphic hardware. Furthermore, such systems overlap the domain of distributed processors such as GPUs and fall outside the neuromorphic scope.
4. Multiplication: Multipliers are typically inefficient to implement in hardware and are limited in standard digital solutions such as Field Programmable Gate Arrays (FPGAs) and Digital Signal Processors (DSPs). Their computational inefficiency and their limited number available on a hardware platform result in neural networks implemented with time-multiplexing. This, in turn, limits the size and the applications where this hardware is viable [20][57].

# Methods and Materials

The elements of SKAN and its learning rule are defined in the first part of this section. In the second part, the dynamical behaviors of SKAN are described.

## SKAN building blocks

At the single neuron level, SKAN consists of a combined synapto-dendritic kernel adaptation and a homeostatic soma with an adapting threshold as shown in Figure 4.

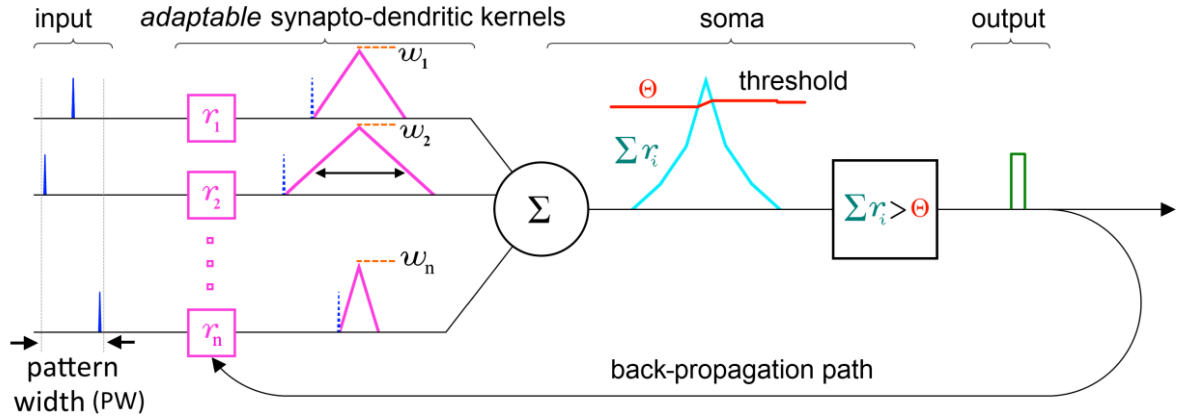

**Figure 4. Schematic of the elements and information paths in a SKAN neuron.** The input spikes (blue) trigger adaptable synapto-dendritic kernels (magenta) which are summed to form the neuron's somatic membrane potential (cyan). This is then compared to an adaptive somatic threshold (red) which, if exceeded, results in an output pulse (green). The output pulse also feeds back to adapt the kernels. Note that in this paper the synaptic weights (orange) are held constant and equal for all synapses. Also note that the back propagating signal does not travel beyond the synapto-dendritic structures of the neuron to previous neural layers.

**Synapse/Dendrite:**

An incoming input spike initiates a simplified synapto-dendritic kernel at each input channel $i$. This kernel is controlled by a physiological process, $p_i$, and for simplicity is modeled as a ramp up and a ramp down sequence generated via an accumulator $r_i$ with step size $\Delta r_i$. An input spike triggers $p_i$, starting the first phase where the accumulator ramps up at each time step $\Delta t$ by $\Delta r_i$ until it reaches a maximum value $w_i$ which represents the synaptic weight, and which is kept constant throughout this paper to simplify the algorithm. After $r_i$ reaches $w_i$, the process switches from the ramp up phase, $p_i=1$, to a ramp down phase, $p_i=-1$, which causes the accumulator to count down at each time step towards zero with the same step size $\Delta r_i$, until it reaches zero, turning off the physiological process, $p_i=0$. It will stay in this state until a new incoming spike re-initiates the sequence. This simple conceptual sequence, which is analogous to a dendritically filtered neuronal EPSP, is illustrated in Figure 5.

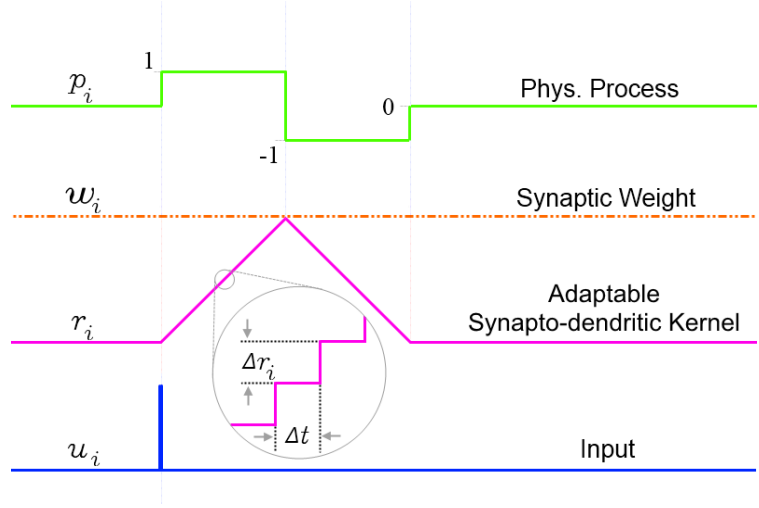

**Figure 5. The simplified adaptable synapto-dendritic kernel of SKAN.** An input spike (blue) triggers the kernel's ramp up ramp down sequence. The input spike sets a flag $p_i$ representing a physical process to one (green). The flag causes an accumulator (magenta) to count up from zero by $\Delta r_i$ at each time step until it reaches $w_i$ (constant orange dotted line), after which the flag is set to negative one, which causes the accumulator to count back down to zero, at which point the flag returns to zero completing the sequence. The value of the accumulator represents the synapto-dendritic kernel, i.e., the post-synaptic potential, which travels to the soma and is summed with other kernels to produce the somatic membrane potential.

The state of the ramp up ramp down flag sequence is described by Equation 1:

**Equation 1:**

$$p_i(t) = \begin{cases} 1 & \text{if } \left(u_i(t) = 1 \wedge p_i(t-1) = 0\right) \vee \left(p_i(t-1) = 1 \wedge r_i(t-1) < w_i\right) \\ -1 & \text{if } \left(p_i(t-1) = 1 \wedge r_i(t-1) \geqslant w_i\right) \vee \left(p_i(t-1) = -1 \wedge r_i(t-1) > 0\right) \\ 0 & \text{else} \end{cases}$$

The *w* parameter in SKAN has similarities to the weight by which a synaptic kernel is multiplied in standard synaptic STDP neuron models and neuromorphic circuits, but with the advantage of not requiring any multipliers, which are otherwise required at every synapse in hardware implementations. The adjustment of *w* in SKAN, via standard synaptic STDP schemes would allow synaptic prioritization and/or the closing off of inactive or noisy channels. The combined effects of dendritic structure and synaptic weight plasticity has only recently begun to be explored, but early evidence points to significant computational power of such a combined system [58]. In this paper, however, in order to clearly demonstrate the stand-alone capabilities of SKAN's synapto-dendritic kernel adaptation mechanism, the synaptic weight parameter of *w* is held constant and is identical for all synapses.

**Soma:**

At the soma the synapto-dendritic kernels are summed together. This summed term is analogous to the membrane potential of a biological neuron. Along with the membrane potential the soma uses a dynamic threshold voltage parameter $\Theta(t)$ and as long as the membrane potential exceeds threshold, the soma spikes, setting the binary *s(t)* from 0 to 1 as described in Equation 2:

**Equation 2**

$$s(t) = \begin{cases} 1 & \text{if } \sum_i r_i(t) > \Theta_{(t-1)} \\ 0 & \text{else} \end{cases}$$

SKAN differs from most previous spiking neuron models in not resetting the membrane potential after spiking (see [12][59] for exceptions). This permits wide pulse widths at the neuron output $s(t)$. While such wide pulses do not resemble the canonical form of the single spike, they are analogous to concentrated spike bursts and play a significant part in the functioning of SKAN.

## Feedback mechanisms/Learning rules:

### Synapto-dendritic kernel slope adaption

One of the central elements of SKAN is the feedback effect of the output pulse $s(t)$ on each of the synapto-dendritic kernels. Here $s(t)$ is analogous to the back propagating spike signal in biological neurons which travels back up the dendrites toward the synapses and is responsible for synaptic STDP.

The logic of the kernel adaptation rule is simple; if a particular dendrite is in the ramp up phase $p_i = 1$ and the back propagation signal $s(t)$ is active, the soma has spiked and this particular kernel is late to reach its peak, meaning that the other kernels have cooperatively forced the membrane potential above the threshold while this kernel has yet to reach its maximum value $w_i$. In response, the ramp's step size $\Delta r_i$ is increased by some small positive value $ddr$ for as long as the output pulse is high ($s(t)=1$) and the kernel is in the ramp up phase. Similarly if a kernel is in the ramp down phase $p_i = -1$ when the back propagation signal is high, then the kernel peaked too early, having reached $w_i$ and ramping down before the neuron's other kernels. In this case the ramp step size $\Delta r_i$ is decreased by $ddr$. Equation 3 describes this simple kernel adaptation rule:

**Equation 3**

$$\begin{bmatrix} r_i(t) \\ \Delta r_i(t) \end{bmatrix} = \begin{bmatrix} r_i(t-1) \\ \Delta r_i(t-1) \end{bmatrix} + p_i(t-1) \begin{bmatrix} \Delta r_i(t-1) \\ ddr \times s(t-1) \end{bmatrix}$$

The use of indirect evidence about the dynamic state of other dendrites in the form of the back propagating spike is a central feature in the operation of SKAN and enables the synchronization of all the neuron's dendritic kernel peaks as shown in Figure 6.

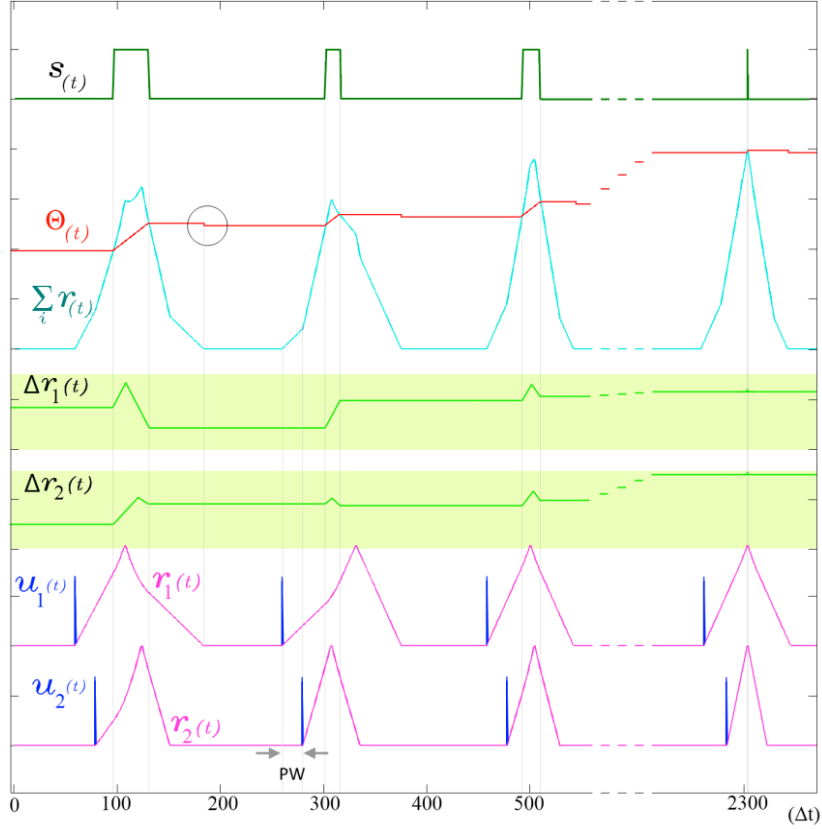

**Figure 6. The adaptation of SKAN.** The kernels and the threshold of SKAN adapt in response to repeated spatio-temporal pattern presentations. For visual clarity the pattern only consists of the Inter-Spike Interval (ISI) across two input channels $u_i(t)$ such that the pattern width (PW) is equivalent to the ISI. By the third presentation of the pattern the kernels have captured the ISI information. With each subsequent presentation the threshold $\Theta(t)$ increases making the neuron more selective as the kernel step sizes $\Delta r_i(t)$ increase making the kernels narrower. As a result each pattern presentation increases the neuron's confidence about the underlying process producing the ISI's, narrowing the neuron's receptive field around the target ISI and producing a smaller output pulse $s(t)$ until, by the 11$^{th}$ presentation ($t$=2300 $\Delta t$), the $\Theta_{rise}$ during the output spike and $\Theta_{fall}$ balance each other such that the $\Theta_{before} \approx \Theta_{after}$. The soma output spike $s(t)$ is now a finely tuned unit delta pulse which indicates high certainty. When the membrane potential returns to zero, the neuron's threshold falls as indicated by the grey circle.

**Threshold adaptation**

The threshold of SKAN is adaptive and changes under two conditions: when the neuron outputs a spike and when the membrane potential returns to zero.

At every time step during an output pulse $s(t)=1$ the threshold increases by $\Theta_{rise}$. This increase in the threshold is analogous to the frequency adaptation effect seen in neurons, which creates a feedback loop reducing the ability of the neuron to spike. Similarly in SKAN, the higher threshold reduces the likelihood and duration of an output pulse. This effect is shown in Figure 6 and described in the first line of Equation 4.

**Equation 4**

$$\Theta(t) = \begin{cases} \Theta(t-1) + \Theta_{rise} & \text{if } \sum_i r_i(t) > \Theta(t-1) \\ \Theta(t-1) - \Theta_{fall} & \text{if } \sum_i r_i(t) = 0 \wedge \sum_i r_i(t-1) > 0 \\ \Theta(t-1) & \text{else} \end{cases}$$

The post spike decrease in threshold $\Theta_{fall}$ operates in opposition to the $\Theta_{rise}$ term. The returning of the membrane potential $\Sigma r_i(t)$ to zero causes a decrease in the threshold by $\Theta_{fall}$ as described by the second line of Equation 4 and shown in Figure 6. The counter balancing effect produced by the $\Theta_{fall}$ and $\Theta_{rise}$ in SKAN is a highly simplified version of the complex mechanisms underlying spike-threshold and frequency adaption in biological neurons [60][61], where excited neurons eventually reach an equilibrium state through homeostatic processes such that the average spike frequency of neurons with a constant input tends asymptotically toward a non-zero value as $t\rightarrow\infty$.

This simple rule set describes all the elements of a single SKAN.

## Single SKAN Dynamics

In this section the dynamics emerging from SKAN's rule are discussed for the single neuron case.

### Observing the first spike in a spike train or burst

As described in the first line of Equation 1 the ramp up phase of the kernel at channel $i$ is only initiated if a spike arrives at the channel ($u_i=1$) while and the kernel is inactive ($p_i=0$). As a result while the $i$th kernel is active no further input spikes are observed. This has the effect that for each input channel the neuron trains on the first spike of a spike train or burst. For the case where the spike train or burst is of shorter duration than the total duration of the kernel, the behavior of the neuron is identical one where the burst is replaced by a single input spike arriving at the start of the burst. The effect of more general Poisson noise spikes is described later in this section.

### Selecting to learn the commonest spatio-temporal patterns

As a single neuron, SKAN has previously been shown to select and learn the most common spatio-pattern presented in a random sequence containing multiple patterns [62]. This effect has been demonstrated in the context of visual processing where hand gestures were transformed to spatio-temporal patterns via a neuronal transform operation [63] and processed by SKAN [64]. Figure 7 shows the performance of a four input neuron as a function of spatio-temporal pattern probability. The graph shows that the neuron's selection of commonest pattern is significantly above chance such that for sequences with $P(x)>0.85$ only the more common pattern will selected.

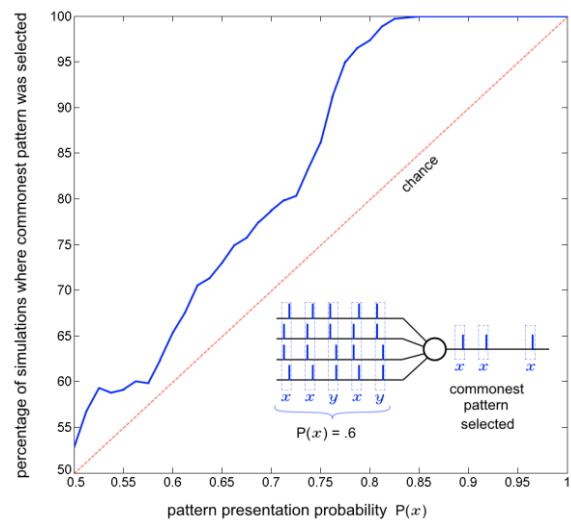

**Figure 7. Commonest pattern selection as a function of pattern presentation probability.** The inset illustrates one simulation a 5 pattern long sequence where each pattern is sampled from two randomly initialized spatio-temporal patterns $x$ and $y$, with probability $P(x)=0.6$. In this particular simulation pattern $x$ was selected by the neuron. The plot shows data resulting from the same experiment but with 1000 simulations of 300 pattern long sequences for each probability $P(x)=0.5$ to 1. The graph shows that the likelihood of a pattern being selected rises with increasing presentation probability. For each simulation the output of the neuron for the second half of the sequence (150th - 300th pattern) was recorded and it was determined whether pattern *x* or pattern *y* had been selected. Also tested was whether both, or neither pattern was selected by the neuron at some point during the sequence (i.e., the neuron spiked at least once for both of the patterns or failed to spike for a pattern it had selected during the sequence). In the more than seven million pattern presentations (1000×150×51) neither of these occurred.

## SKAN response time improves with adaptation without information loss

In addition to the kernel adaptation and increasing threshold effect, the response time of SKAN, i.e., the time from the last arriving input spike in a pattern to the neuron's output spike, decreases with every pattern presentation. This effect, shown in Figure 8, is absent in the standard STDP schemes where improved response times comes at the cost of information loss. In STDP schemes the earliest spikes in a spatio-temporal pattern tend to be highly weighted while the later spike lose weight and have little effect on recognition [84]. This behavior can be seen as advantageous if an assumption is made that the later spikes in carry less information however in this is an assumption that cannot be made in general. In contrast SKAN's adaptable kernels reduce output spike latency with adaptation while still enabling every spike to affect the output. This effect proves critical in the context of a multi-SKAN competitive network, where the best-adapted neuron is also always the fastest neuron to spike.

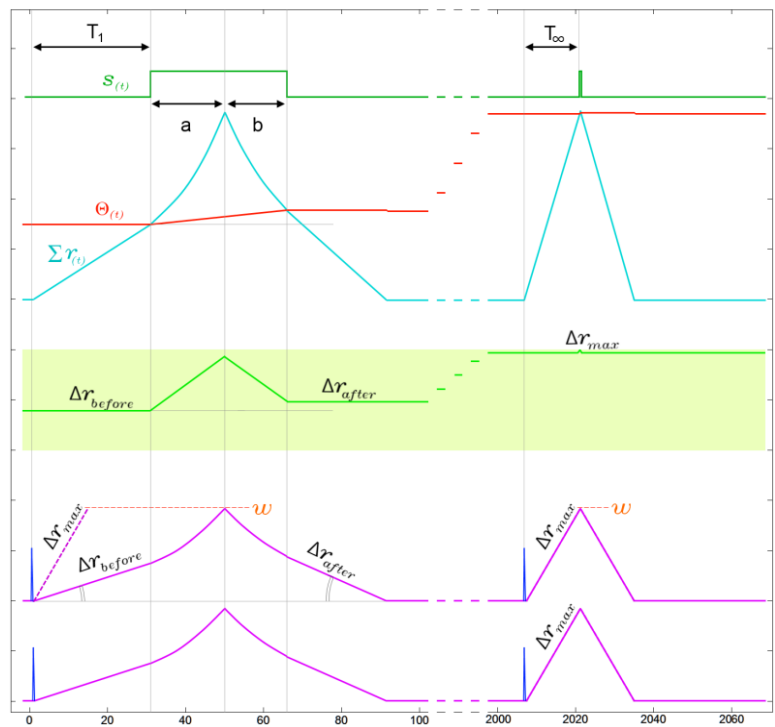

**Figure 8. Narrowing of kernels leads to improved response time during neuronal adaptation in a two input neuron.** For visual clarity the neuron is presented with an ISI=0$\Delta t$ pattern and the two kernels start with identical initial slopes ($\Delta r_1(0) = \Delta r_2(0) = \Delta r_{before}$). In the region under the output pulse, $r(t)$ is the second integral of the constant *ddr* and therefore follows time symmetric parabolic paths (a) and (b) as it rises and falls. However due to the threshold rise which also occurs during the output pulse, the output pulse is not symmetric around the $r(t)$ peak, such that the parabolic ramp down phase (b) is shorter than the parabolic ramp up phase (a). As a result of this asymmetry $\Delta r_{after}$ is larger than $\Delta r_{before}$. This effect increases the kernel's slope $\Delta r$ with each pattern presentation, narrowing the kernels until $\Delta r$ reaches $\Delta r_{max}$. As a result of this narrowing, the response time of the neuron from last arriving input spike to the rising edge of the output, which is $T_1$ in the first presentation, improves until it reaches its minimal possible value $T_\infty \approx w / \Delta r_{max}$.

As shown in Figure 8, the combination of the kernel and threshold adaptation rules of SKAN increases $\Delta r$ and decreases the response time between the last arriving input spike and the rising edge of the output spike with each presentation. If this increase is left unchecked $\Delta r$ will increase until it equals $w$ at which point the kernels take the shape of a single pulse such that $T_\infty = 1\Delta t$. To prevent this $\Delta r$ must saturate at $\Delta r_{max}$ as shown in Figure 8 with $\Delta r_{max}$ limited by Equation 5

. This restriction ensures that the kernel of the first spike in an input pattern cannot return to zero before the last spike in the pattern arrives enabling all kernels to converge due to feedback from the same output signal.

**Equation 5**

$$\Delta r_{max} < w / PW$$

where $PW$ is the maximal pattern width of the target pattern.

**Evolution of the temporal receptive field in SKAN approximates statistical inference**

Recent work has demonstrated the connection between synaptic weight adaptation and approximate probabilistic inference in the context of rate coding and spiking networks [65][66][67][68][69][70][71][72], where typically the state of binary hidden variables are inferred from noisy observations using a large number of neurons. In this section we show that synapto-dendritic kernel adaptation enables a single neuron to make statistical inferences not about binary hidden variables but about hidden ISI generating processes. Figure 9 illustrates the evolution of the temporal receptive field of a neuron with two inputs as the neuron attempts to learn the statistics of an underlying process that produces ISIs with linearly increasingly temporal jitter. The receptive field of the neuron describes the amount by which the membrane potential $\Sigma r_i(t)$ exceeds the threshold $\Theta(t)$ as a function of the input spike pattern times of $u_i(t)$. For the simple two input case illustrated, the receptive field is a scalar function of the one-dimensional ISI. In order to calculate the receptive field, following each pattern presentation the neuron's new parameters ($\Delta r_i$ and $\Theta$) were saved and the neuron was simulated repeatedly using these saved parameters for every possible ISI given the maximum pattern width $PW$. For each simulation the summation in Equation 6

 was calculated at the end of the simulation resulting in the receptive fields shown in Figure 9.

**Equation 6**

$$RF_{i=2}(\tau) = \sum_t \left( \sum_i r_i(\tau,t) - \Theta(\tau,t) \right) \times s(\tau,t)$$

where $\tau$ is the ISI being simulated.

The ISI at which the receptive field expression above is at its maximum (RF Max) indicates the ISI for which the neuron is most receptive and may be interpreted as the ISI expected by the neuron. Similarly the ISI boundary where the receptive field expression goes to zero is the limit to the range of ISI's expected by the neuron. An ISI falling outside the receptive field boundaries results in no spike and no adaptation but simply reduces the neuron's confidence and can be viewed as outlier.

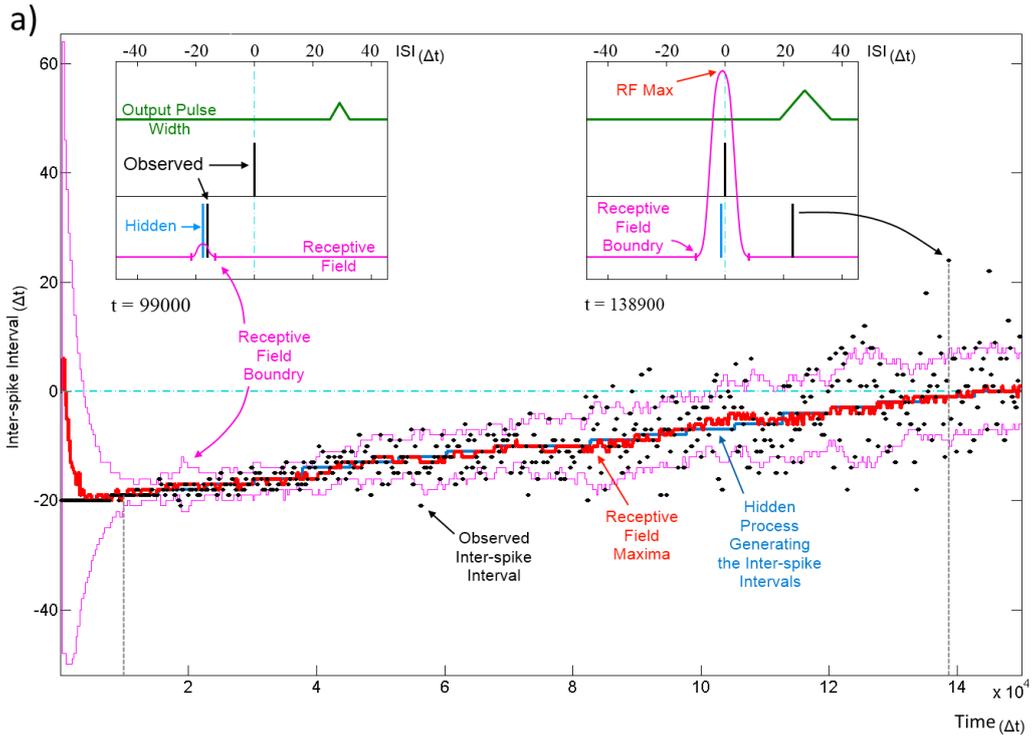
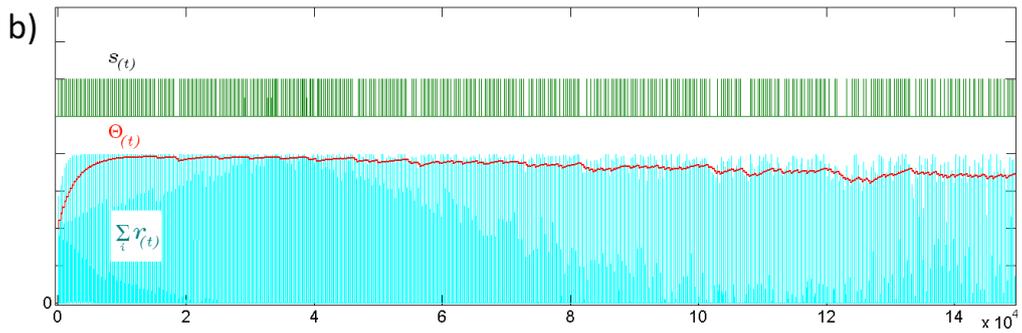
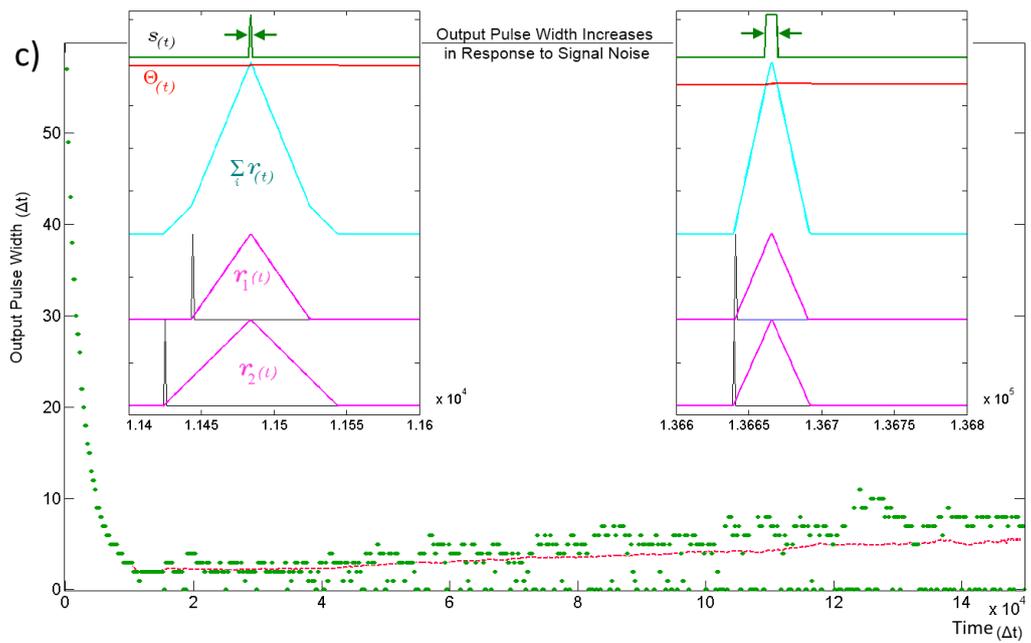

**Figure 9. Tracking a hidden ISI producing process and its variance.** All three panels (a, b and c) show different aspects of the same simulation where a single SKAN learns statistics of a dynamic ISI across two input channels. a) A hidden process (blue) moves from ISI=-20 $\Delta t$ to ISI=0 $\Delta t$. The process begins with no temporal jitter noise, such that the observed ISI's (black dots) equal the hidden process ($\sigma$=0 $\Delta t$) and the blue hidden process is covered by the observed black dots. At $t$=0, the sum of the neuron's randomly initialized kernels peak at ISI=6 $\Delta t$. As the kernels adapt and capture the ISI information, the receptive field maximum (red line) converges on to the observed ISIs. This causes the threshold to rise shrinking the receptive field to a minimum size (left inset $t$ = 9.9 k$\Delta t$). At this stage receptive field boundaries (pink lines) lie very close on either side of the hidden process' mean value. As the simulation continues the noise in the ISI producing process is deliberately increased linearly with time. The neuron continues to follow the process but with every spike that doesn't land on the receptive field maximum the threshold falls slightly which increases the receptive field size and increases the neuron's receptivity to ever more unlikely observations (right inset $t$=138.9 k$\Delta t$). b) Variables and behavior of the neuron throughout the simulation: after a rapid initial increase the threshold $\Theta(t)$ settles near the peak of the membrane potential $\Sigma r_i(t)$. As the noise increases linearly the threshold begins to fall gradually. Missing output spikes in the $s(t)$ spike train correspond to input spikes that have landed outside the receptive field boundaries. c) The output pulse width (green = observed, red = running average) increases with increasing signal noise. As with the receptive field size, the output pulse width is initially large. As more patterns are observed, the threshold rises and settles just below the peak of the membrane potential and the pulse width reaches a minimum width of 1-2 $\Delta t$. At this low noise level ($\sigma \approx 0.5 \Delta t$) there are no missing outputs, such that all pulse widths are above zero. As the noise increases, more ISIs land away from the receptive field maximum and some fall completely outside the receptive field, decreasing the threshold, which results in wider output pulse width whenever observed ISI's *do* land near the receptive field maximum. The dashed magenta line tracks the mean spike width, which also increases with noise. This illustrates that the mean output pulse width of SKAN is a reliable correlate of input noise level.

Figure 9 a) shows SKAN's receptive fields tracking the statistics of a moving ISI generating process with dynamic noise levels with a high level of accuracy such that the blue line indicating the hidden process is barely visible from under the red line marking the receptive field maximum. Figure 9 c) shows the neuron transmitting wider output or bursts with increasing noise. In addition, increasing ISI noise causes a growing gap between the envelope of the pulse widths and the running average of the pulse widths. This increasing gap is critical to the operation of the neuron, as it is caused by missed pattern presentations, i.e., patterns that produce no output pulse because of the presented noisy pattern being too dissimilar to the one the neuron has learnt and expects. The effect of a missed pattern is a fall in the neuron's threshold by $\Theta_{fall}$. When presented with noiseless patterns this fall would be balanced almost exactly by the threshold rise due to the $\Theta_{rise}$ term in Equation 4 during the output pulse. However, without the output spike there is a net drop in threshold. Yet this lower threshold also makes the neuron more receptive to noisier patterns creating a feedback system with two opposing tendencies which:

1. Progressively narrows kernels around the observed input pattern while shrinking the neuron's receptive field by raising the threshold.
2. Expands the receptive field in response to missed patterns by reducing the threshold while allowing the kernels to learn by incorporating ever less likely patterns.

The balance between these two opposing tendencies is determined by the ratio $\Theta_{rise}$:$\Theta_{fall}$, which controls how responsive the neuron is to changing statistics. With a stable noise level SKAN's dynamics always move toward an equilibrium state where the neuron's tendency to contract its receptive field is precisely balanced by the number of noisy patterns not falling at the receptive field maximum. This heuristic strategy results in the receptive field's maximum and extent tracking the expected value of the input ISI's and their variance respectively as shown in Figure 10.

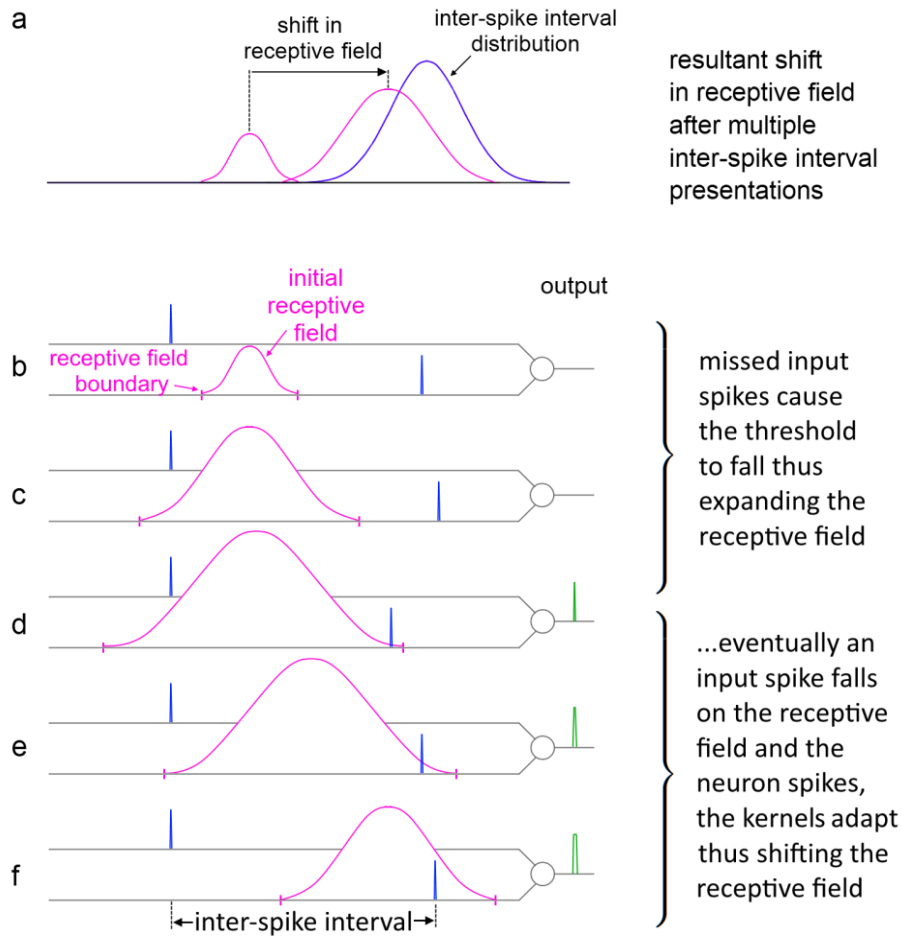

**Figure 10. Evolution of SKAN's receptive field in response to input.** a) Total resultant change in SKAN's receptive field after multiple pattern presentations. b) SKAN with a small initial receptive field which does not match the ISI distribution. The input spike lands outside the receptive field boundaries. c) As more ISI's fall outside the small receptive field the threshold falls and the receptive field expands, but without shifting the position of its maximum value. d) An ISI just falls on to the greatly expanded receptive field producing an output spike. e) The output spike causes the SKAN kernels to adapt shifting the receptive field toward the true position of the underlying process. f) As more and more ISI's fall closer to the receptive field maximum wider output pulses are produced which adapt the kernels faster shifting the receptive field more rapidly while the resultant rise in the threshold contracts the receptive field. With enough observations the receptive field would eventually become centered on the input ISI distribution with the receptive field boundaries tracking the ISI's distribution.

## Learning in the presence of Poisson spike noise and missing target spikes

In addition to robustness to temporal jitter in the put pattern an important feature of neural systems is their performance in the presence of Poisson spike noise. Recent work has highlighted that unlike most engineered systems where noise is assumed to degrade performance, biological neural networks can often utilize such noise as a resource [73][74][75]. In the neuromorphic context the performance of neural network architectures in the presence of noise is well documented [76][77][78]. To test SKAN's potential performance in stochastic real world environments, the combined effects of extra noise spikes as well as missing target spikes needs to be tested. Figure 11 illustrates how different signal to noise ratios can affect SKAN's ability to learn an embedded spatio-temporal spike pattern.

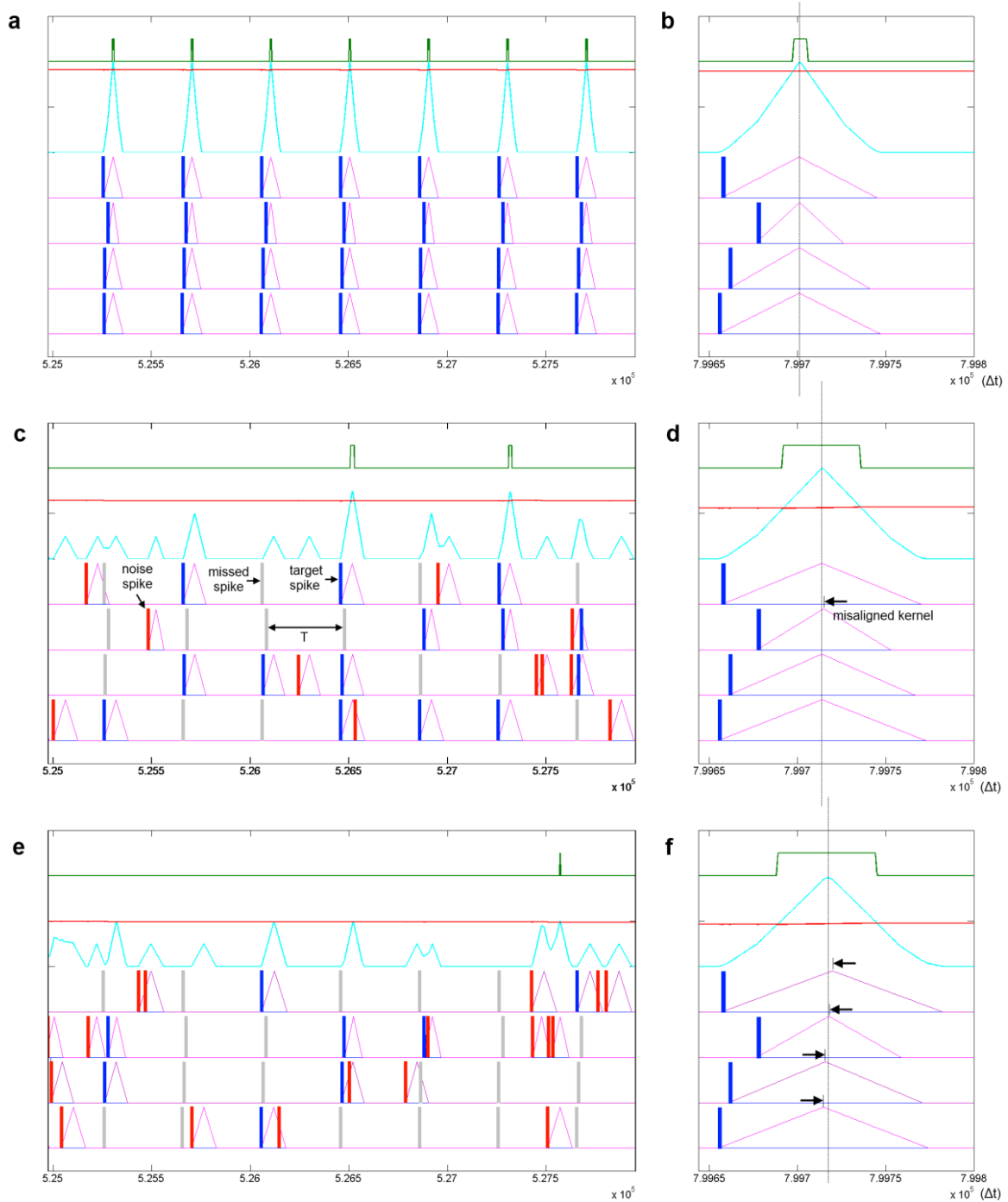

**Figure 11. Learning spatio-temporal spike patterns in the presence of both Poisson spike noise and missing target spikes.** Panel (a) shows the presentation of seven patterns in the middle of a simulation sequence with a noiseless environment. The kernels are highly adapted ($\Delta r_2 = \Delta r_{max}$), the threshold is high and the output spikes are narrow indicating high certainty. Panel (b) shows the result of a final noiseless test pattern at the end of the simulation showing in detail that the kernels resulting from the test pattern peak at the same time. Panel (c) shows the same interval of the same simulation as panel (a) but with a 1:1 signal to noise ratio where the probability of a target spike being deleted is half or P(signal)=0.5 and the Poisson rate is also half such that P(noise) = 0.5/T. Panel (d) shows the result of a noiseless test pattern presentation at the end of the simulation. The increased level of noise has resulted in an incorrect ramp step ($\Delta r_2$) such that the $r_2$ kernel peaks slightly late (black arrow). Panel (e) shows a simulation with a 1:2 signal to noise ratio. Panel (f) shows that the high noise level has resulted in slight misalignment of all four kernels.

To quantify the performance of SKAN in the presence of Poisson noise and missing target spikes a series of simulations each comprising of two thousand pattern presentations were performed. At the end of each simulation the RMS error between the neuron's receptive field maxima and the random target pattern was measured and is shown in Figure 12.

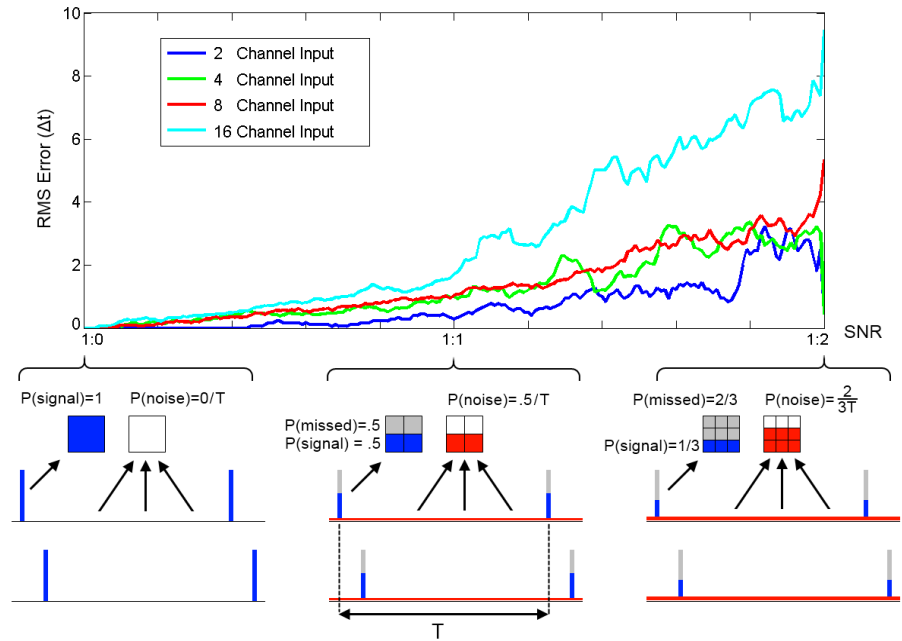

**Figure 12. RMS error between receptive field maxima and target spike patterns as a function of spike signal to noise ratio.** The three bottom panels show the spike probability distributions at three points along the SNR axis. The signal spikes (blue), missed spikes (grey), and noise spikes (red) are illustrated for the three cases of 1:0, 1:1 and 1:2 signal to noise ratios. The mean spike rate was maintained at 1 spike per channel per time period between pattern presentations T. At the completion of a simulation with one thousand pattern presentations the RMS error between the resulting receptive field maxima and the target spatio-temporal pattern was calculated. As the plot illustrates the error increases with noise and simulations of neurons with more input channels resulted in higher error.

## Multi-SKAN classifier

In order to extend a single learning neuron to a classifier network it is important that different neurons learn different patterns. Ideally a neuron in a layer should not be in anyway affected by the presentation of a pattern that another neuron in the same layer has already learnt or is better placed to learn.

As outlined in Equation 3, SKAN adapts its kernels only during an output pulse. This rule is particularly conducive to competitive learning such that the simple disabling of the neuron's spiking ability disables all learning. Whereas previously proposed algorithms utilize multi neuron Winner-Take-All layers with real valued rate based inhibitory signals to prevent correlated spiking and maximize the network learning capacity [79][80], in a SKAN network a simple global inhibitory OR gate serves the same function. The reason a simple binary signal can be used here is that in a SKAN network the best-placed neuron for any pattern will be the fastest neuron to spike. This allows a layer of neurons with shared inputs to learn to recognize mutually exclusive spatio-temporal patterns. To this end, Equation 2, describing the neuron's output, is replaced by Equation 7 (underlined terms added). The addition of a global decaying inhibitory signal as described in Equation 8, act on all neurons to disable any rising edge at the output. This means that neuron *n* can only *initiate* an output spike $s_n$ if no other neuron has recently spiked, i.e., the inhibitory signal is inactive ($inh(t-1)=0$) and it can only *continue* spiking if it was already spiking in the last time step ($s_n(t-1)=0$).

**Equation 7**

$$s_n(t) = \begin{cases} 1 & \text{if } \sum_i r_{n,i}(t) > \Theta_{n\,(t-1)} \land \underline{\left(inh_{(t-1)} = 0 \lor s_{n\,(t-1)} = 1\right)} \\ 0 & \text{else} \end{cases}$$

**Equation 8**

$$inh_{(t)} = \begin{cases} inh_{max} & \text{if } \bigcup_n s_n(t) = 1 \\ inh_{(t-1)} - inh_{decay} & \text{if } inh_{(t-1)} > 0 \\ 0 & \text{else} \end{cases}$$

As shown in Figure 13 and described in Equation 8, the inhibitory signal is realized via an OR operation on the output of all neurons, and a decaying behavior which keeps the inhibitory signal active for a period of time after a neuron has spiked to prevent spiking by other neurons. After the output spike ends, this feedback loop decays from $inh_{max}$ by $inh_{decay}$ at each time step until reaching zero at which point the global inhibitory signal turns off allowing any neuron to spike. As shown in Figure 13 the decay only begins at the end of the pulse making the inhibitory signal operate as a global peak detector which stays at $inh_{max}$ for the duration of the pulse, ensuring that the inhibitory signal robustly suppresses spiking activity for a wide range of potential output pulse widths.

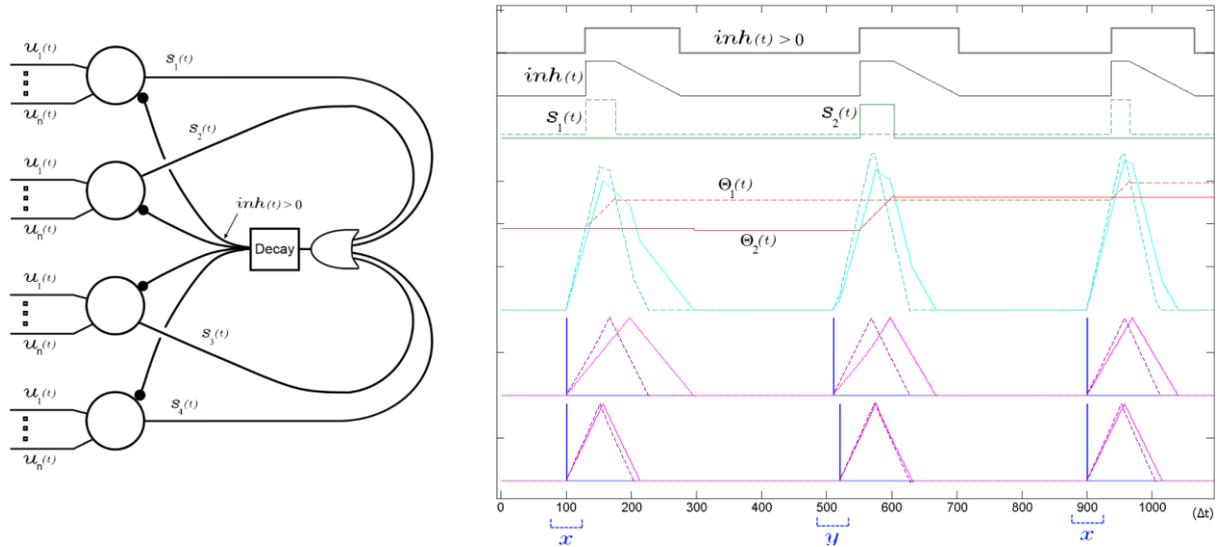

**Figure 13. A single global decaying inhibitory signal suffices to push apart the neurons' receptive fields and decorrelate the spiking of the SKAN network.** Left panel shows the network diagram of four neurons with an inhibitory signal. The decay feedback loop extends the duration of the inhibitory signal beyond the initial triggering spike via the $inh(t)$ signal using a counter and a comparator in the decay block. The right panel shows the simulation results from a two input two neuron network learning to classify two ISI's $x$ and $y$. The sum of the randomly initialized kernels of neuron one (dashed) happen to peak earlier than neuron two (solid) so that neuron one fires first in response to the first pattern ($x$ with ISI=0 $\Delta t$). During this first output pulse neuron one's threshold rises sharply reducing its receptivity, while its kernel step sizes adapt towards each other such that $\Delta r_{1,1} \approx \Delta r_{1,2}$. Meanwhile the inhibitory signal blocks neuron two from spiking when its kernel sum exceeds its threshold only a few time steps after neuron one, which means the neuron is prevented from adapting to pattern $x$. At the second pattern presentation pattern $y$ is shown (ISI=10 $\Delta t$). For this pattern the sum of the kernels of the second neuron, still unchanged from their random initialization, reach that neuron's threshold slightly earlier than neuron one and so neuron two spikes and begins adapting to pattern $y$. A subsequent presentation of pattern $x$ again triggers neuron one and the kernels of the two neurons increasingly fine tune to their respectively chosen pattern with each presentation as their thresholds rises reducing their receptivity to other patterns.

As with the single neuron output rule, the single neuron threshold adaptation rule of Equation 4 can be modified to Equation 9 (underlined terms added) to utilize the global inhibitory signal for the multi-neuron case. This modification prevents a neuron's threshold being affected by the presentation of patterns that another neuron is better adapted to. The addition of the underlined terms in the first line of Equation 9 means that a neuron's threshold can only rise when its membrane potential exceeds its threshold and the inhibitory signal is not already active, or if the neuron itself spiked in the previous time step. The fall in the threshold is similarly conditioned on the neuron having spiked *before* the global inhibitory signal was activated, such that only the very best adapted neuron, i.e., the one that generated the inhibitory signal in the first place, adapts its threshold.

**Equation 9**

$$\Theta_n(t) = \begin{cases} \Theta(t-1) + \Theta_{rise} & \text{if } \sum_i r_i(t) > \Theta(t-1) \wedge \left(inh(t-1) = 0 \vee s_n(t-1) = 1\right) \\ \Theta(t-1) - \Theta_{fall} & \text{if } \left(\sum_i r_i(t) = 0 \wedge \sum_i r_i(t-1) > 0 \wedge inh(t-1) = 0\right) \vee \left(s(t) = 0 \wedge s(t-1) = 1\right) \\ \Theta(t-1) & \text{else} \end{cases}$$

Such a global inhibitory signal has been utilized in LIF neurons [81][82] and synaptic weight STDP neurons as a means of decorrelating neuronal firing patterns [83][84]. Here, however, its use is subtly different from both. Although in LIF and synaptic STDP architectures and in SKAN a global inhibitory signal results in the decorrelation of output spikes, in the purely synaptic weight adapting schemes the neuron's response time remains static and does not improve with adaptation and in the LIF networks [81][82] there is no lasting adaptation at all. SKAN's improved response time due to kernel adaptation and the global inhibitory signal realize a positive feedback mechanism absent in previous models. In a SKAN network a neuron's small initial advantage for a pattern results in a slightly earlier output spike. This output spike globally inhibits all other neurons, which in turn results in exclusive adaptation to the pattern by the first spiking neuron. This further improves that neuron's response time for the pattern and increases the likelihood of the neuron being the first to spike due to a subsequent presentation of the same pattern, even in the presence of temporal jitter. Thus, the adaptation of SKAN's kernels and thresholds, together with the global inhibitory network, mean that the neuron whose initial state is closest to the presented pattern will be the first to respond and prevent all other neurons adapting to this pattern. This effectively "hides" the pattern from the other neurons and allows unsupervised spike pattern classification by the network as whole as demonstrated in the proceeding results sections.

There are two important constraints adhered to by the preceding modification of the SKAN rules. The first constraint is that the required connectivity does not increase combinatorially with the number of neurons as described in Equation 10 since the only feedback path is from the single global inhibitory signal.

**Equation 10**

$$\text{total connections} = (\text{number of input channel} + 2) \times \text{number of neurons}$$

The second constraint is that no complex central controller is required for arbitration between the neurons. In competitive neural network schemes where a neuron's fitness is expressed as a real value from each neuron to a Winner-Take-All network, multiple bits (connections in hardware) are required to transport this information. Alternatively rate based systems encode such real valued signal over time in their spike rate which are then utilized by a corresponding rate based Winner-Take-All system. But in SKAN these requirements are reduced. Since a neuron's latencies correlates with its adaptation to a target pattern, the neurons do not need to report a real value but only a single bit. This mode of operation can be interpreted as either a connectivity saving or as a speed saving with respect to alternative multi-bit or rate based systems respectively. Furthermore because of the robustness of the system, checking for, or prevention of, simultaneous output spikes is not necessary. Random initial heterogeneities in the neurons' parameters and/or noise in their signals is enough to eliminate the need for central control by pushing the neurons away from input space saddle points toward their stable non-overlapping receptive fields.

# Results

## Online unsupervised spatio-temporal spike pattern classification

In the following sections the classification performance of SKAN is tested in several ways. For these tests equally likely spatio-temporal spike patterns, each with one spike per channel per presentation were presented in random sequences to the SKAN network. Table 1 details the parameters used in all the tests. These parameters were deliberately chosen for non-optimized performance so as to try to mimic the use of the system in the wild by a non-expert user. Examples of available optimizations include: higher *ddr* values which result in faster converging systems, reduced $\Theta_{rise}/\Theta_{fall}$ ratio for improved robustness to noise, increased $\Delta r_{max}/\Delta r_{i,n}(t=0)$ ratio and increased pattern widths for enhanced pattern selectivity.

| Parameter | Value | Description |
|---|---|---|
| $ddr$ | 1 | Change in the kernel step size. Higher value results in faster adaptation; lower values are more robust to noise. |
| $w(r_{max})$ | 10000 | Maximum kernel height (Synaptic weight). |
| $r_{min}$ | 0 | The kernel signal $r(t)$ saturates at zero. |
| $\Delta r_{i,n}(t=0)$ | 100×(1+ rand) | Initial kernel step size (For each input *i* to each neuron *n*). The randomized initialization allows different neurons to learn different patterns. |
| $\Delta r_{max}$ | 400 | Maximum kernel step size. |
| $r_{i,n}(t=0)$ | 0 | Initial kernel value. |
| $\Theta_{rise}$ | 40 × *inputs* | Rise in threshold during output spike, where *inputs* is the number of input channels per neuron. |
| $\Theta_{fall}$ | 100× *inputs* | Fall in threshold due to input spikes, where *inputs* is the number of input channels per neuron. |
| $inh_{max}$ | 100 | Initial value of the inhibitory countdown . |
| $inh_{decay}$ | 1 | Step size of the inhibitory countdown. As a rule of thumb use: $inh_{max}/inh_{decay}= \min(\Delta r_{i,n}(t=0))$. |
| T | 400 $\Delta t$ | Time between pattern presentations |

**Table 1 Parameter values used for all results.**

## Hardware efficiency through 1-to-1 neuron to pattern allocation at the local level

Through temporal competition a local network of mutually inhibiting SKANs can efficiently distribute limited neural resources in a hardware implementation to observed spatio-temporal patterns as is demonstrated in Figure 14.

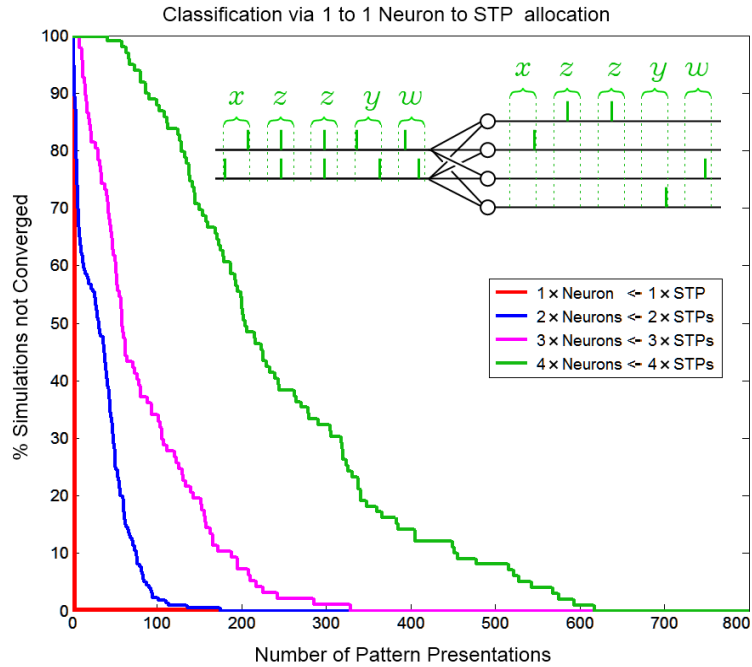

**Figure 14. Convergence rate of as a function of neuron/pattern numbers and number of pattern presentations for a 1-to-1 two input neuron to pattern allocating network.** As the number of patterns/neurons increases the system requires longer pattern sequences to correctly allocate exactly one unique pattern to each neuron. The inset shows the five consecutive correct classifications of four patterns by four neurons.

Similar to biological systems, in a SKAN network there is no supervisor switching the network from a training mode to a testing mode so there is no distinction between learning and recognition. This means that attempting to test SKANs in the traditional neural network sense by switching off a network's adaptation mechanisms would disable the system. Thus to test the network's performance 1000 simulations were generated for each instance of the network, with up to 800 pattern presentations each. The network was considered to have converged to a stable solution when twenty consecutive patterns were correctly classified by the network, i.e., with a single neuron responding per spatiotemporal pattern. This is illustrated in the inset of Figure 14. Correct classification was defined as the case where a neuron spikes if and only if its target pattern is presented and where the neurons consistently spike for the same learnt target pattern. Also, a single neuron should spike once for each input pattern and no extra output spikes occur. The percentage of simulations that had not converged to correct classification was recorded as a function of the number of patterns presented, and is shown in Figure 14. Simulations were terminated once a network had converged. The number of consecutive patterns was chosen as twenty to reduce the likelihood that the observed "correct" response of the network was due to chance.

**Classification performance as a function of spatio-temporal pattern dimension.**
The problem of coordinating multiple synapses for unsupervised neuronal classification in spiking neural network models, whether through simply learning synaptic weights or through more complex pathways, is difficult [85]. In SKAN the hybrid synapto-dendritic kernel adaptation produces convergence profiles shown in Figure 15. These results show how the convergence profiles of SKAN change with the number of active input channels. Additionally, the right panel in Figure 15 shows the effect of increasing the resolution of the spatio-temporal pattern. Doubling the number of time steps in the maximal width of the target pattern *PW*, results in improved convergence profiles.

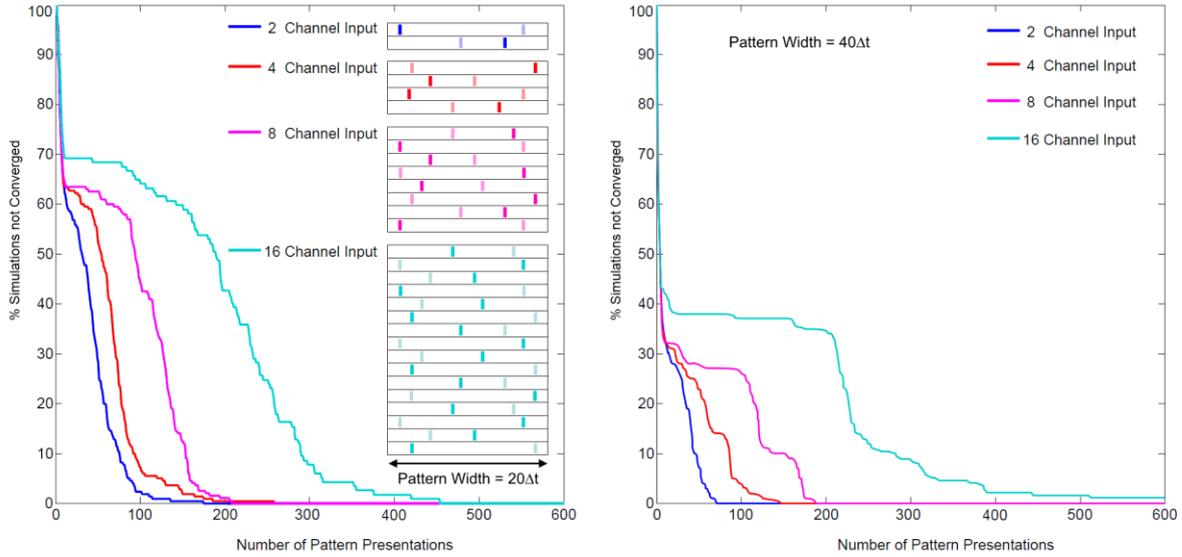

**Figure 15. Convergence rates as a function of input channel dimension and pattern width.** Left panel: two random target patterns (light and dark bars) of maximal pattern width $PW = 20\ \Delta t$ and of dimensions 2, 4, 8 and 16, were presented at random to a two neuron network, with the convergence of simulations plotted over the number of presentations. Right panel: the same test with maximal pattern width $PW = 40\ \Delta t$.

**Classification in the presence of temporal noise**

In order for SKAN to operate as an effective classifier competing neurons must balance the requirements of selectivity and generalization. In the spatio-temporal context, generalization takes the form of temporal jitter noise. In this context neurons must recognize patterns closest to their learnt target pattern despite the presence of temporal noise, while not recognizing other similarly noise corrupted patterns that are closer to the target patterns learnt by other neurons. Furthermore, the neurons should not expect the learning phase to be any less noisy than the testing phase or even for there to be any such distinct separation between learning and recognition. As well the neurons should maintain their correct learning and recognition behavior across a wide range of noise levels and they should ideally do so without the requirement for external adjustment of their parameters. SKAN satisfies all these requirements. The classification performance of SKAN is robust to temporal jitter noise as illustrated in Figure 16 where two neurons act as two Kalman filters with shared inputs attempting to learn the statistics of two noisy but distinct ISI generating processes.

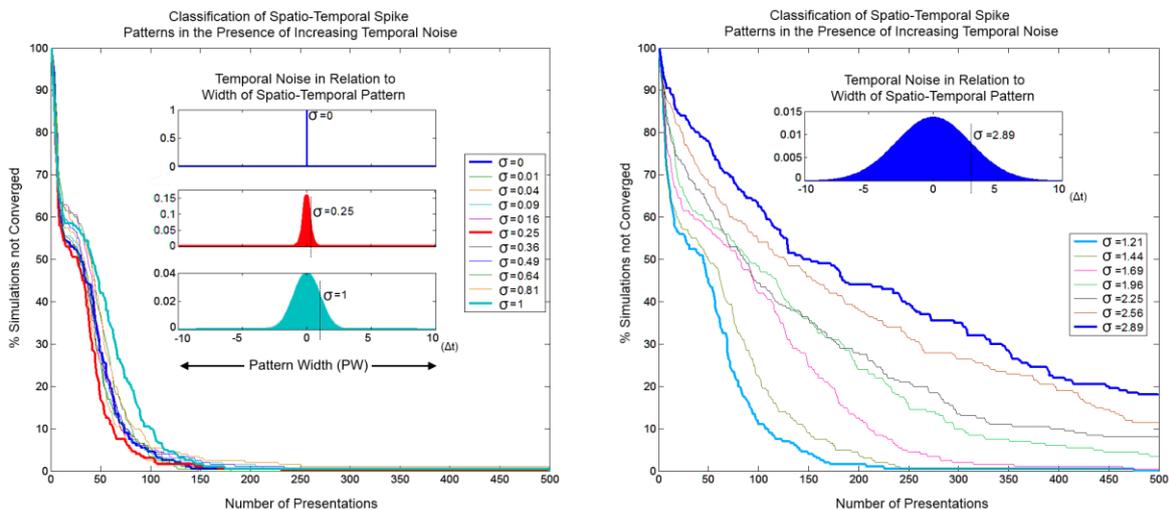

**Figure 16. Convergence as a function of temporal noise and pattern presentations in a two neuron network with two input channels.** The insets illustrate selected noise distributions relative to the maximal pattern width (PW = 20 $\Delta t$). The left panel shows convergence profiles due to temporal noise distribution with standard deviation σ = 0-1 $\Delta t$. At lower noise levels the convergence profile is approximately the same or faster (red) than the zero noise case. The right panel shows the same for σ ≈ 1-3 $\Delta t$.

Because of the constant adaptation of the neurons, moderate levels of temporal noise with standard deviation up to σ =0.25 $\Delta t$, which is 1/80th of the pattern width, either do not affect or actually *improve* SKAN performance. With high temporal noise levels, i.e., with a standard deviation that is 1/20th the width of the pattern (σ=1 $\Delta t$), the convergence profile is still similar to that of the noiseless case. Such levels of temporal noise can disable a conventional processor and even some neural networks. Even at the extreme, with noise that has a standard deviation more than 1/7th of the pattern width, some simulations still result in the neurons correctly classifying the separate ISI sources.

As a temporal coding scheme, the robustness of SKAN's learning algorithm to temporal noise is critical for potential real-world applications, where the ability to operate (and degrade gracefully) in noisy, dynamic environments is favored over ideal performance in ideal noise free circumstances.

## Implementation in FPGA

SKAN was implemented in an Altera Cyclone-V GX FPGA, a low-end FPGA containing 77,000 programmable logic elements (LEs). The functions of SKAN were programmed based on the equations described in the earlier sections, written using the Verilog hardware description language, with no optimization techniques employed. A key feature of this design is that no multipliers are required: SKAN is executed entirely using simple summation and logical operations only, thus significantly reducing computational complexity and hardware resources. Registers are used to store required design parameters. Table 2 shows the utilization of the FPGA in terms of registers, adaptive logic modules (ALMs) and the percentage of resources used for SKAN modules containing different number of synapses. From this we can see that SKAN is efficient in its usage of hardware resources. Results from the FPGA are identical to the simulated results as integers were used for both the simulations and the FPGA and therefore no approximations were required. Integers were used to avoid floating point operations, thereby reducing computation. An efficient use of hardware resources, reduced computational effort, and its ease of implementation make SKAN an attractive neuromorphic solution in terms of both cost and performance.

|               | FPGA Resource Usage       |                                    |                  |
| :-----------: | :-----------------------: | :--------------------------------: | :--------------: |
| **No. of Synapses** | **Single bit Registers** | **Adaptive Logic Modules (ALMs)**[*] | **Usage Percentage** |
| 1  | 48  | 189  | 0.6% |
| 2  | 72  | 297  | 1.0% |
| 4  | 121 | 501  | 1.7% |
| 8  | 218 | 922  | 3.2% |
| 16 | 411 | 1580 | 5.4% |

[*] An ALM is equivalent to 2.65 logic elements (LEs)

**Table 2 Altera Cyclone V FPGA resource usage for a SKAN neuron with different number of synapses.**

## Discussion

As outlined in the introduction a limiting factor in many neuromorphic systems is the large number of complex synapses which require multipliers and high connectivity networks required for robust performance. A simple solution to this challenge has been to physically implement of one or a few instances of these complex elements and use time multiplexing and AER to generate larger virtual networks. The kernels of SKAN which do not require multipliers allow more synapses to be physically realized in hardware while their adaptability means that better performance can be achieved using fewer synapses. Furthermore the time based operation of the neurons reduces the required connectivity. This potentially allows entire networks to be physically implemented in hardware. Such small or medium sized networks whose behaviors have been described in this report can then be cascaded or multiplexed to form larger networks. Such solutions could potentially occupy a middle ground between fully hardware implemented networks with limited connectivity but high bandwidth and single neuron realizations with high connectivity and limited operating speeds. While the focus of this introductory report is on characterization of small non-optimized SKAN networks, preliminary work on the application of the architecture to larger, more difficult recognition tasks such as unsupervised learning of the MNIST dataset has not revealed any limits to the capabilities of larger, more optimized networks. Future work will focus on comparison of SKAN networks to other neural network solutions on established datasets, comparison of the inference capabilities of the neuron to optimal probabilistic estimators and the investigation of the combined effects of adaptation of SKAN's kernels and the adaptation of its synaptic weight parameter *w* which allows encoding of synaptic signal to noise ratios for each input channel.

## Conclusion

In this paper we have presented the Synapto-dendritic Kernel Adapting Neuron – SKAN, a neuromorphic implementation of a spiking neuron that performs statistical inference and unsupervised learning and spatio-temporal spike pattern classification. The use of simple adaptable kernels was shown to represent an efficient solution to hardware realized neural networks without the need for multipliers while SKAN operation was shown to be robust in the presence of noise allowing potential applications in noisy real-world environments. Finally it was shown that SKAN is hardware efficient and easily implemented on an FPGA.